# Joint Learning-based Causal Relation Extraction from Biomedical Literature

Dongling Li[1], Pengchao Wu[1], Yuehu Dong[1], Jinghang Gu[2], Longhua Qian[1*], Guodong Zhou[1]

[1]School of Computer Science and Technology, Soochow University, Suzhou, Jiangsu Province, China, 215006.

[2]Department of Chinese and Bilingual Studies, The Hong Kong Polytechnic University, Hong Kong, China, 999077.

{20204227045, 20204227037, 20215227045}@stu.suda.edu.cn, gujinghangnlp@gmail.com, {qianlonghua, gdzhou}@suda.edu.cn

**Abstract**

Causal relation extraction of biomedical entities is one of the most complex tasks in biomedical text mining, which involves two kinds of information: entity relations and entity functions. One feasible approach is to take relation extraction and function detection as two independent sub-tasks. However, this separate learning method ignores the intrinsic correlation between them and leads to unsatisfactory performance. In this paper, we propose a joint learning model, which combines entity relation extraction and entity function detection to exploit their commonality and capture their inter-relationship, so as to improve the performance of biomedical causal relation extraction. Meanwhile, during the model training stage, different function types in the loss function are assigned different weights. Specifically, the penalty coefficient for negative function instances increases to effectively improve the precision of function detection. Experimental results on the BioCreative-V Track 4 corpus show that our joint learning model outperforms the separate models in BEL statement extraction, achieving the F1 scores of 58.4% and 37.3% on the test set in Stage 2 and Stage 1 evaluations, respectively. This demonstrates that our joint learning system reaches the state-of-the-art performance in Stage 2 compared with other systems.

## 1 Introduction

With the rapid development of biomedical research, the volume of biomedical literature is growing consistently. When dealing with such large-scale textual data, it is extremely difficult to manually mine useful information from them because of intensive labor and prohibitive cost. Therefore, researchers pay more and more attention to how to automatically obtain relevant knowledge from biomedical literature efficiently and effectively, which leads to the emergence of biomedical text mining research. Biomedical text mining aims to extract useful information from biomedical literature and transform it into structured knowledge to enrich the content of the domain-specific knowledge base. Concurrently, the structured representation of biomedical knowledge is also one of the hot spots pursued by field experts due to its application in knowledge graph construction, knowledge reasoning, and completion. Some well-known attempts for biomedical entity networks are the system biology markup language (SBML) [1], biological pathway exchange language (BioPAX) [2], and biological expression language (BEL) [3].

  The BioCreative-V community [4] organized the shared task 4, which aims to extract causal relations between biomedical entities from literature and express them in the BEL form. BEL statements that represent causal relations are not only suitable for machine processing but also easily readable for humans. They can express not only causal relations between entities but also relations between entity functions. Therefore, BEL statements have a rich representation ability for biomedical knowledge. Consequently, they pose new challenges to biomedical text mining research owing to their complex structures. So far, the methods used to extract BEL statements can be roughly divided into two groups: event-centered and entity-centered.

Thanks to many publicly available biomedical event corpora, the event-centered approach aims to map biomedical events extracted from literature to entity functions and relations and then formulate them into BEL statements. Ravikumar et al., 2017 [5] propose to extract entity and event information through a rule-based semantic analyzer [6] and then combine them into a complete BEL statement. The disadvantages of rule-based methods are low coverage and poor generalization ability to other domains. Choi et al., 2016 [7] use the event extraction system of machine learning-based TEES [10] to extract biomedical events and then convert them into entity relations and functions in BEL statements. They also discuss the impact of entity co-reference resolution on the extraction performance of BEL statements. Although there is much similarity between BEL statements and biomedical events [13], task difference between them is non-trivial, leading to unsatisfactory performance for BEL statements.

The entity-centered approach takes entities as the core component, and directly determines entity functions and causal relations between entities, and then formulates them into BEL statements. In the NCU-IISR system, Lai et al., 2016 [8] firstly detect the functions of entities through keyword matching, and secondly use the biomedical semantic role annotation method [9] to parse sentences into a predicate-argument structure (PAS), and then convert them into Subject-Verb-Object (SVO) triplets to extract causal relations between entities. Finally, entity relations and functions are combined into BEL statements. This method ignores the sentences without SVO structure, leading to low extraction recall. Liu et al., 2019 [11] propose to divide the BEL extraction task into two sub-tasks: entity relation extraction (RE) and entity function detection (FD). They use two separate BiLSTM models based on the attention mechanism to extract entity relations and detect entity functions during the BEL statement construction. Following their work, Shao et al., 2021 [12] further propose a BEL statement extraction method based on Simplified Biological Expression Language (SBEL), which acts as an intermediate form between BEL statements and entity relations/functions to retain relation and function instances as many as possible. Moreover, BERT models are applied to improve the performance of relation extraction and function detection due to their language representation superiority. However, their treatment of relation extraction and function detection as two independent sub-tasks ignores their commonality and correlations. As suggested by Liu et al., 2019 [11] that the precision of function detection dominates the contribution of the function detection to the BEL extraction, low precision in function detection deteriorates the overall performance in BEL statement extraction. Therefore, they simply use a threshold filtering approach to find more accurate entity functions to formulate the BEL statements.

BEL statement extraction involves two sub-tasks: entity relation extraction and entity function detection, it thus is essentially taken as a multi-task learning (MTL) problem. For deep learning models, MTL generally improves learning performance by hard or soft parameters sharing in neural networks. The hard parameter sharing means sharing some hidden layers of all tasks in order to mitigate the risk of overfitting during training (Baxter et al., 1997) [15]. They either merely share the word embeddings (Collobert et al., 2008) [19], or share the whole sentence encoder (Liu et al., 2019 [18]), or generate labels of different tasks at different levels of the neural networks (Søgaard et al., 2016 [24], Sanh et al., 2019 [14]). Soft parameter sharing means that each sub-task has its own model with respective parameters, and the similarity of some parameters is guaranteed by regularizing the distance metric between model parameters. The regularization method can be the L2 norm (Duong et al., 2015 [16]) or the trace norm (Yang et al., 2016 [17]).

Motivated by the success of multi-task learning in NLP and SBEL-based causality extraction [12], we propose a joint learning model of relation extraction and function detection based on hard parameter sharing [18] for BEL statement extraction. These two sub-tasks share the same BERT encoder but with different fully connected layers and output layers. The joint model is trained with a novel joint loss function. Specifically, different function types in the loss function are assigned with different weights, which improves the function detection precision by penalizing the negative instances and further promotes the overall extraction performance for BEL statements.

## 2 SBEL

Following the work by Shao et al., 2021 [12], SBEL statements are adopted as the bridge between BEL statements and learning instances. The basic idea is to convert BEL statements on the training set into SBEL statements for model training, and then conduct the SBEL statement predictions on the test set and further assemble the predicted results into BEL statements. SBEL simplifies the BEL extraction task while retaining most of the information in BEL statements, which effectively improves the BEL extraction performance. It is assumed that a causal relation only exits between a pair of entities, and each entity has at most one function. Particularly, for the function type *complex* that has more than one entity, the instance is accordingly decomposed into multiple function instances, and thus new multiple SBEL statements are formed subsequently. Formally, an SBEL statement can be expressed in the following five-tuple form:

$$\text{SBEL:} \quad <\textit{func1} \ \text{em1} \ \textit{relation} \ \textit{func2} \ \text{em2}>$$

Where *em1* and *em2* are two entity mentions indicating subject and object, respectively. *func1* and *func2* are the corresponding functions, e.g., *act*, *pmod*, etc. The *relation* represents the causal relation between the two entities, e.g., *increases* and *decreases*.

Figure 1 illustrates an example sentence (SEN: 10021786) with BEL and SBEL statements as well as the learning instances in the separate models and the joint model, respectively. There are two BEL statements (BEL:20038346 and BEL:20038344) among three protein entities (T2: ITK, T1: IL-5 and T3: IL-3). These two BEL statements are converted into SBEL1 and SBEL2 statements, respectively. For simplicity, the *kin* function type of ITK is generalized to the function type *act*.

Shao et al., 2021 [12] decompose these two SBEL statements into two relation instances (ReInst1, ReInst2) and three function instances (FdInst1, FdInst2, FdInst3), and use two independent models to address these two sub-tasks. The disadvantage of the separate models is that they cannot consider the inter-relationship between two sub-tasks. Another limitation called one-entity-one-function is that an entity mention can only have one function type, which is not consistent with the definition of BEL statements. For example, the entity ITK has the *act* function in SBEL1 but *None* in SBEL2, however, when function instances are generated in separate models, only one function type is retained, here *act* for the protein mention T2.

Different from their work [12], our approach is to address the sub-tasks of relation extraction and function detection simultaneously by training one joint model directly with SBEL instances (SbelInst1 and SbelInst2 in Figure 1). In this way, we not only consider the inter-relationship between two sub-tasks (e.g., when the relation between two entities does not exist, their functions cannot exist) but also remove the limitation of one-entity-one-function. In other words, an entity mention may have different function labels in different SBEL statements, e.g., both *act* in SbelInst1 and *None* in SbelInst2 are applied to the same protein mention T2.

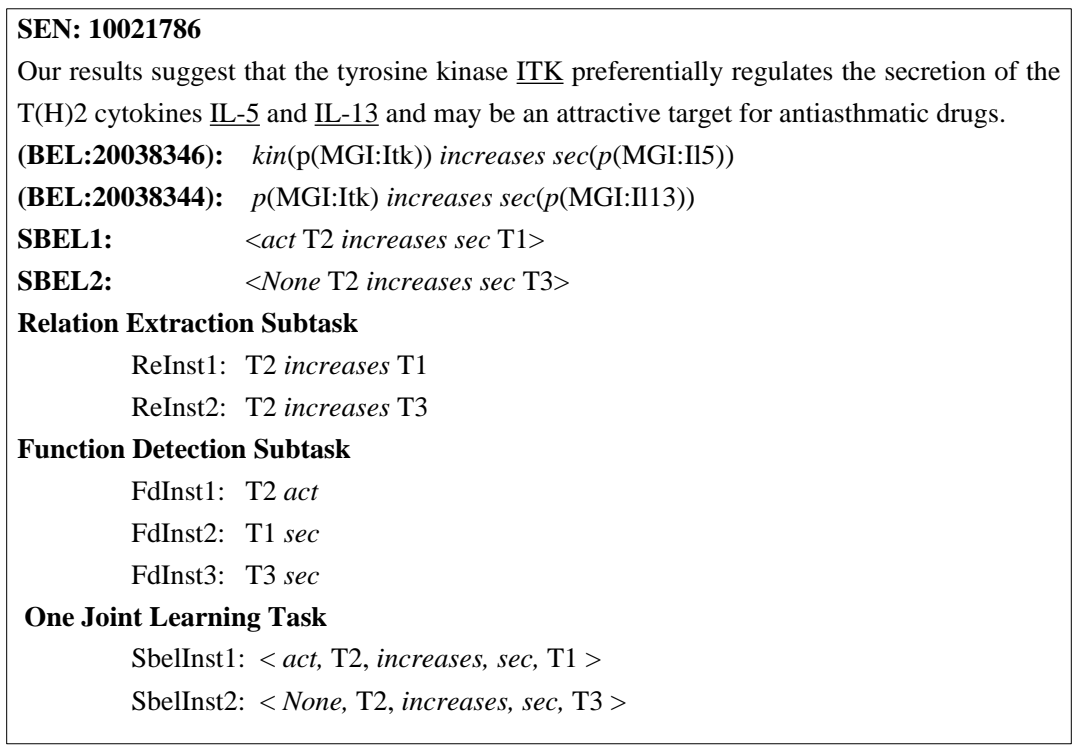

**SEN: 10021786**

Our results suggest that the tyrosine kinase ITK preferentially regulates the secretion of the T(H)2 cytokines IL-5 and IL-13 and may be an attractive target for antiasthmatic drugs.

**(BEL:20038346):** *kin*(p(MGI:Itk)) *increases sec*(p(MGI:Il5))

**(BEL:20038344):** *p*(MGI:Itk) *increases sec*(p(MGI:Il13))

**SBEL1:** <*act* T2 *increases sec* T1>

**SBEL2:** <*None* T2 *increases sec* T3>

**Relation Extraction Subtask**
    ReInst1: T2 *increases* T1
    ReInst2: T2 *increases* T3

**Function Detection Subtask**
    FdInst1: T2 *act*
    FdInst2: T1 *sec*
    FdInst3: T3 *sec*

**One Joint Learning Task**
    SbelInst1: < *act,* T2, *increases, sec,* T1 >
    SbelInst2: < *None,* T2, *increases, sec,* T3 >

Figure 1. Learning instances of separate models and joint model

## 3 Methodology

### 3.1 Framework

The framework of causal relation extraction based on joint learning is shown in Figure 2. Different from the separate models (Shao et al., 2021 [12]), it is not necessary to decompose SBEL into relation instances and function instances. On the contrary, SBEL instances are directly used to train a joint-learning model. In testing, the relation and functions of a pair of entity mentions are predicted simultaneously. The main steps are as follows:

    (1) **Conversion**: BEL statements are converted into corresponding SBEL instances. This step is the same as Shao et al. 2021 [12].

    (2) **Training**: SBEL instances in the training set are used to train the joint learning model. The next subsection will describe this step.

    (3) **Prediction**: For each pair of entities, the relation between them and the function of each entity are predicted simultaneously. If there exists a relation between them, the predicted SBEL statements are assembled into the corresponding BEL statements.

### 3.2 Joint learning of SBEL extraction

Due to the great success of the pre-trained language model in natural language processing, e.g., BERT [20], and the derived BioBERT [21] in relation extraction in biomedical text mining, we adopt the BioBERT model as the encoder for joint learning. BERT is a masked language model using the transformer [22] as a feature extractor. The self-attention mechanism enables the BERT model to have a broader global view, so it can better capture sentential semantic information.

The pre-training corpus of BERT is BooksCorpus [23] and English Wikipedia datasets. However, biomedical literature contains a large amount of domain knowledge, and the general domain pre-training language model may not perform well in biomedical medical mining. Therefore, we adopt the BioBERT pre-trained model [21], which uses PubMed abstract and PMC full text as the pre-training datasets. It has achieved excellent performance in many text mining tasks in the biomedical field. Figure 3 shows the model diagram of joint learning based on BioBERT, which combines relation extraction and function detection. We describe the input/output layers and the training loss function as follows.

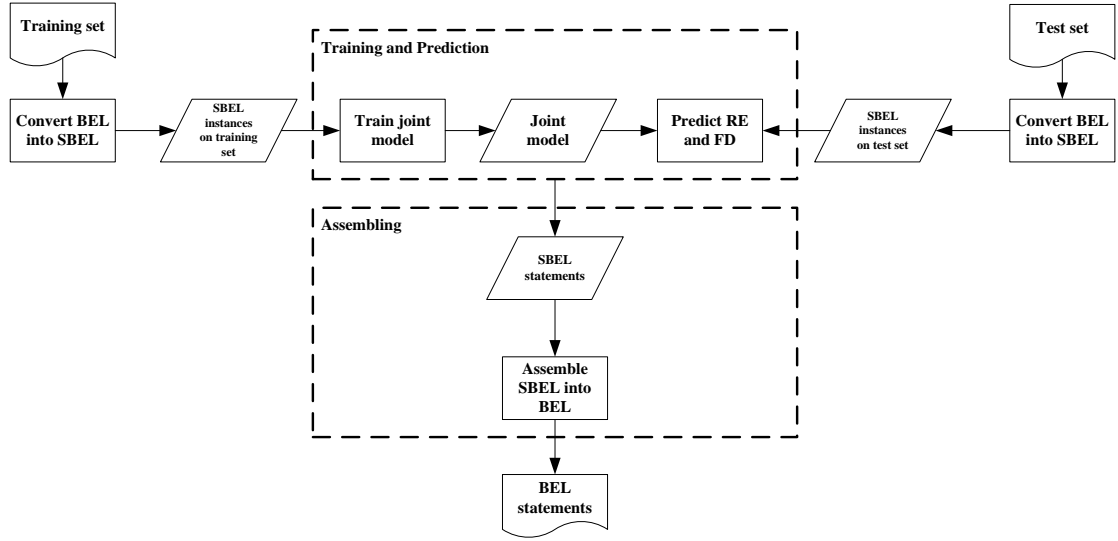

Figure 2. Framework of joint extraction of BEL statements based on SBEL statements.

### 3.2.1 Input Layer

Given a sentence $S = \{w_1, w, \ldots, w_M\}$, where $w_i$ is the i-th word and $M$ is the number of words in the sentence. The sentence is tokenized into word pieces and three special tokens are inserted at the beginning to generate a sequence of tokens $Tokens = \{[CLS], [F1], [F2], tok_1, tok_2, \ldots, tok_N\}$ as the model input, where $N$ denotes the number of tokens in the sentence. [CLS] is used as the classification token for relation extraction, [F1] and [F2] are used as the classification tokens for function detection of two entities respectively. Unlike [CLS], [F1] and [F2] are not inherent tokens in the BERT vocabulary, so we replace them with two reserved symbols [unused1] and [unused2] in the BERT vocabulary. And $E = \{E_{[CLS]}, E_{[F1]}, E_{[F2]}, E_1, E_2 \ldots E_N\}$ is the sequence of token embeddings. Similar to BioBERT [21], two special symbols '@' and '$' are inserted around entities to mark subject and object respectively.

### 3.2.2 Output Layer

Each token in $Tokens$ is encoded by BERT to generate the hidden representation sequence $H = \{C, f1, f2, H_i, H_{i+1} \ldots H_N\}$. $C, f1, f2$ are passed to two fully connected layers and three *softmax* classifiers to predict the relation type in the relation label set $L_r = \{l_1, l_2, \ldots, l_{n_r}\}$ and function types in the function label set $L_f = \{l_1, l_2, \ldots, l_{n_f}\}$ respectively, i.e., $[\hat{y}_r, \hat{y}_{f1}, \hat{y}_{f2}]$. They can be formulated as follows:

$$\hat{y}_r = softmax(W_r C + b_r) \quad (1)$$
$$\hat{y}_{f1} = softmax(W_f f1 + b_f) \quad (2)$$
$$\hat{y}_{f2} = softmax(W_f f2 + b_f) \quad (3)$$

Where $W_r \in \mathbb{R}^{d_z \times n_r}, W_f \in \mathbb{R}^{d_z \times n_f}$, $d_z$ is the dimension of the BERT embeddings, $n_r = 3$ is the number of relation types, $n_f = 7$ is the number of function types, $b_r, b_f$ are the bias. Note that the fully connected layer of entity 1 and entity 2 functions shares the same set of parameters $[W_f, b_f]$.

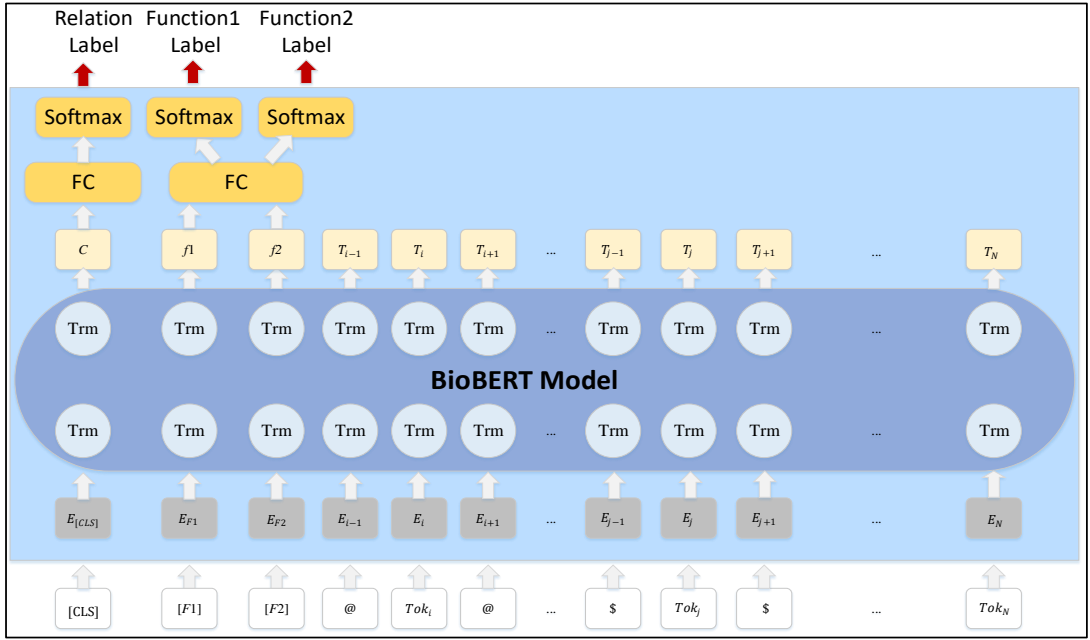

Figure 3. Joint learning model based on BERT.

### 3.2.3 Loss function

The loss function of the joint model is the sum of the cross-entropy loss of relation, entity 1 function, and entity 2 function, as shown in formula (4). As the previous research work [11, 12] points out, the precision of entity function detection plays an important role in the performance of BEL statement extraction. Therefore, we assign different weights to different function types in the loss of entity functions. Specifically, the more the penalty coefficient of the *None* function type, the more the improvement in the precision of entity function detection. The model training adopts the Adam optimization algorithm to optimize in the direction of loss minimization. The training loss can be formalized as below:

$$J = -\left(\sum_{i=1}^{N}\sum_{j=1}^{n_r} y_{j,r} log\hat{y}_{j,r} + \sum_{i=1}^{N}\sum_{j=1}^{n_f} w_j y_{j,f1} log\hat{y}_{j,f1} + \sum_{i=1}^{N}\sum_{j=1}^{n_f} w_j y_{j,f2} log\hat{y}_{j,f2}\right) \quad (4)$$

Here, $N$ is the length of the input token sequence, $y_{j,r}$ is the gold relation label, $y_{j,f1}, y_{j,f2}$ are the gold function labels for entity 1 and entity 2, respectively. $w_j \in \mathbb{R}^{n_f}$ represents the weight of the j-th function type.

## 4 Experimentation

### 4.1 Datasets

The corpus provided by the BioCreative-V BEL task includes one training set and one test set, where each sentence is annotated with one or more BEL statements. Table 1 reports the statistics on sentences, BEL statements, SBEL statements, and relations and functions derived from SBEL statements in our joint model (denoted as JNT) and separate models (denoted as SEP) in Shao et al., 2021 [12]. The mark "-" in a cell indicates that the number in the joint model is the same as that in the separate models. It can be seen from the table that:

Table 1. BC-V BEL corpus statistics

| Statistics | Training | | Test | |
|---|---|---|---|---|
| | SEP[11] | JNT | SEP[11] | JNT |
| Sentence | 6,353 | - | 105 | - |
| BEL | 11,066 | - | 202 | - |
| SBEL | 10,097 | - | 203 | - |
| Relations | 10,097 | - | 203 | - |
| *increases* | 7,382 | - | 150 | - |
| *decreases* | 2,715 | - | 53 | - |
| Functions | 6,476 | 7,759 | 69 | 87 |
| *act* | 4,637 | 5,497 | 27 | 32 |
| *deg* | 119 | 137 | 6 | 6 |
| *pmod* | 712 | 832 | 5 | 5 |
| *sec* | 217 | 251 | 4 | 4 |
| *tloc* | 63 | 71 | 5 | 4 |
| *complex* | 728 | 971 | 22 | 36 |

- The numbers of SBEL statements and relation instances are the same because both statistics are acquired based on two entities, i.e., a pair of entities only have one SBEL statement and one relation instance.
- The difference lies in the number of function instances. Generally, there are more function instances in the joint model than in the separate models because of their different statistical methods, i.e., two functions are counted in an SBEL instance while only one function is counted in a function instance.

**4.2 Evaluation**

We adopt commonly-used evaluation metrics to evaluate the extraction performance, namely Precision ($P$), Recall ($R$), and $F1$ measure. Precision refers to the ratio of the number of correct samples to the total number of samples predicted by the model. Recall refers to the ratio of the number of correct samples predicted by the model to the number of all gold samples. $F1$ is the harmonic mean of precision and recall. The aforementioned measures are calculated as follows:

$$P = \frac{TP}{TP + FP} \quad (5)$$

$$R = \frac{TP}{TP + FN} \quad (6)$$

$$F1 = \frac{2 \times P \times R}{P + R} \quad (7)$$

Where True Positive (TP) and False Positive (FP) represent the numbers of correctly and wrongly predicted positive samples, respectively, while False Negative (FN) represents the number of wrongly predicted negative samples.

Due to its complexity, BEL statement extraction is evaluated at six evaluation levels: Term Level (Term), Function Level (Func), Relation Level (Rel), Statement Level (State), Function Secondary Level

(FS), and Relation Secondary Level (RS). More details of these evaluation levels can be referred to [4].

In order to better evaluate the SBEL statement extraction performance, we also define three additional evaluation levels: SBEL_RE, SBEL_FD, and SBEL. The SBEL_RE level evaluates whether the relation triplet <em1, relation, em2> in the SBEL statement is correct, ignoring their functions. The SBEL_FD level evaluates whether the functions of the subject and object entities (em1 and em2) are correct when a relation holds between them. The SBEL level evaluates whether the whole SBEL statement, i.e., the SBEL five-tuple, is correct. In our joint model, the test instances are all SBEL statements, so the performance at three levels can be calculated directly while in the separate models, we need to assemble the relation between two entities and their functions into SBEL statements, and then evaluate them.

### 4.3 Models for Comparison

Four different models as follows are compared to evaluate the effectiveness of joint learning and function weighting:
- SEP [12]: we rerun the separate models of Shao et al., 2021 [12] as a baseline. The sub-tasks of relation extraction and function detection are trained and applied to prediction separately, and then their results are assembled into SBEL statements and finally BEL statements.
- SEP_CW: similar to SEP except that function weighting is applied to the loss function of the function detection model.
- JNT: our joint model that can simultaneously train relation extraction and function detection. The SBEL statements predicted by the model are directly assembled into BEL statements.
- JNT_CW: similar to JNT except that function weighting is applied to the joint loss function.

### 4.4 Settings

The "biobert-pubmed-v1.1" version of BioBERT is used as the pre-trained model. The hyper-parameters of the joint learning model are shown in Table 2.

Table 2. Hyper-parameters of neural networks

| Hyper-parameters | Value |
| --- | --- |
| batch_size | 16 |
| epoch | 3 |
| max_seq_len | 128 |
| loss function | categorical_crossentropy |
| learning rate | 1e-5 |
| optimizer | Adam |
| dev_ratio | 0.1 |
| funtion_type_weight | [1, 1, 1, 1, 1, 1, 3] |

The best-performing model on the development set is selected, which is further evaluated on the test set.

### 4.5 Experimental Results

**The effect of function weighting in separate models in Stage 2 evaluation**

Table 3 compares the performance of SEP and SEP_CW on the test set in Stage 2 evaluation. There are five evaluation levels, where the levels of RE and FD represent the performance for relation extraction and function detection respectively; the level of SBEL represents the performance of SBEL statements assembled from the relation and function instances; that of State (REL) indicates the performance of BEL statements when only relations are considered; and that of State (MRG) indicates the performance of BEL statements when functions and relations are merged. We take the average performance of five runs as the overall performance. The highest P/R/F1 values in each row are shown in bold, and the values in the bracket on the right of F1 measures represent the standard variances across five runs. As can be seen from the table:

- Since the function weighting only affects the function detection, the RE and State (REL) performance of the two models are the same, and the precision of FD is improved with nearly 12 units by SEP_CW. Although the recall and F1 scores of FD decrease, the P/R/F1 values of SBEL extraction of SEP_CW are still better than those of SEP. This is probably because the function weighting significantly reduces the number of wrongly recognized entity functions in SBEL statements. Accordingly, SEP_CW also outperforms SEP in recall and F1 scores of State (MRG).

Table 3. Stage 2 performance comparison of SEP and SEP_CW

| Evaluation levels | SEP [12] | | | SEP_CW | | |
|---|---|---|---|---|---|---|
| | P | R | F1 | P | R | F1 |
| RE | 74.7 | 62.0 | 67.8(±1.8) | 74.7 | 62.0 | 67.8(±1.8) |
| FD | 45.5 | **54.5** | **49.5(±5.3)** | **57.4** | 38.0 | 45.2(±2.7) |
| SBEL | 52.5 | 46.4 | 49.3(±1.4) | **53.5** | **47.3** | **50.2(±1.3)** |
| State (REL) | 59.8 | 46.2 | 52.1(±1.2) | 59.8 | 46.2 | 52.1(±1.2) |
| State (MRG) | **64.1** | 49.6 | 55.9(±3.3) | 63.3 | **51.0** | **56.5(±1.2)** |

- The performance stability of the model is also improved by function weighting. The standard variance of F1 scores of FD decreases from 5.3 to 2.7, and the standard variance of F1 performance of State (MRG) decreases as well. It might be because fewer but more accurate positive instances are predicted, leading to the more robust performance of function detection.

**The effect of joint learning in Stage 2 evaluation**

In Table 4, we compare the performance of our joint model JNT with the separate models SEP in Shao et al., 2021 [12] in Stage 2 evaluation. Here, the performance scores of relation extraction and function detection are evaluated on the SBEL_RE and SBEL_FD levels (see Subsection 4.2). The performance of SBEL_RE in Table 4 is the same as RE in Table 3 while that of FD is different. The reason is that the FD performance in Table 3 is evaluated in terms of entity function instances while SBEL_FD is evaluated in terms of entity functions in SBEL statements. Similarly, we take the average of five runs as the overall performance. The higher P/R/F1 scores in JNT are shown in bold, while the P/R/F1 scores in JNT_CW larger than those counterparts in JNT are highlighted and underlined. It can be seen from the table:

- The performance of JNT is better than that of SEP at almost all levels except SBEL_RE. Among them, the F1 score at the State (MRG) level has been improved by 1.1 units.
- As for SBEL_FD, the precision and recall of JNT increase by 8.3 units and 3.5 units respectively compared with SEP, suggesting that JNT can effectively improve the precision of function detection without reducing its recall.
- As for SBEL_RE, the recall and F1 scores of JNT are improved compared with SEP, implying that JNT can not only improve the SBEL_FD performance but also improve the SBEL_RE performance.

Table 4. Stage 2 performance comparison of SEP, JNT and JNT_CW

| Evaluation levels | SEP [12] | | | JNT | | | JNT_CW | | |
|---|---|---|---|---|---|---|---|---|---|
| | P | R | F1 | P | R | F1 | P | R | F1 |
| SBEL_RE | **74.7** | 62.0 | 67.8 | 74.5 | **63.2** | **68.3** | 74.0 | 62.2 | 67.6 |
| SBEL_FD | 57.7 | 25.8 | 35.6 | 66.0 | **29.3** | **40.5** | <u>72.6</u> | 17.0 | 27.5 |
| SBEL | 52.5 | 46.4 | 49.3 | **57.3** | **48.7** | **52.6** | 56.9 | 47.9 | 52.0 |
| State (REL) | 59.8 | 46.2 | 52.1 | **60.9** | **48.3** | **53.9** | 60.9 | 48.3 | 53.9 |
| State (MRG) | 64.1 | 49.6 | 55.9 | 64.5 | 51.2 | 57.0 | <u>66.0</u> | <u>52.4</u> | <u>58.4</u> |

- Similar to function weighting in the separate models, the precision of SBEL_FD is significantly improved by function weighting in the joint model. Although the recall is reduced, the overall performance of BEL statements is still improved by 1.4 units of F1 score.

**Performance comparison of four models in Stage 1 evaluation**

Gold entities in the test set are given in Stage 2 evaluation while entities must be automatically recognized and normalized in Stage 1 evaluation. We adopt the same method to recognize and normalize entities as [11,12]. Table 5 compares the performance scores of SEP, SEP_CW, JNT, and JNT_CW on the test set in Stage 1 evaluation. The highest P/R/F1 scores in each row are shown in bold. As can be seen from the table:

Table 5. Stage 1 performance comparison of different models

| Evaluation levels | SEP [12] | | | SEP_CW | | | JNT | | | JNT_CW | | |
|---|---|---|---|---|---|---|---|---|---|---|---|---|
| | P | R | F1 | P | R | F1 | P | R | F1 | P | R | F1 |
| SBEL_RE | 47.8 | 56.6 | 51.8 | 47.8 | 56.6 | 51.8 | **49.9** | **60.1** | **54.5** | 48.8 | 58.3 | 53.1 |
| SBEL_FD | 57.9 | 23.2 | 33.0 | 62.1 | 19.4 | 29.4 | 75.1 | **31.3** | **44.1** | **83.3** | 14.5 | 24.7 |
| SBEL | 35.3 | 41.8 | 38.3 | 34.9 | 45.2 | 39.5 | **38.9** | **46.8** | **42.4** | 37.0 | 44.2 | 40.2 |
| State (REL) | 34.6 | 29.8 | 32.0 | 34.8 | **32.6** | **33.6** | **35.3** | 31.2 | 33.1 | 34.8 | 30.7 | 32.6 |
| State (MRG) | 38.0 | 32.6 | 35.0 | 37.4 | 35.1 | 36.2 | **39.6** | **35.3** | **37.3** | 39.6 | 35.2 | 37.2 |

- Compared with the Stage 2 performance, the performance scores of all four models decrease drastically, which is obviously caused by the noise in automatically recognized named entities.
- Whether for the joint models or the separate models, function weighting consistently improves the precision of SBEL_FD. However, in separate models, it improves the overall performance of BEL statements, while in the joint model, it does not.
- The performance scores of two joint models are generally better than those of two separate models, and this improvement is due to that the joint model can greatly improve the precision of SBEL_FD.

**Performance comparison with other systems**

Table 6 compares the performance of our joint model with other systems on the BC-V BEL extraction task, which are rule-based [4,5], event-based [6], and other machine learning methods i.e., NCU-IISR [8], Att-BiLSTM [11], and SBEL-BERT [12]. The upper part of the table shows the performance in Stage 1 while the lower part shows the performance in Stage 2. The highest performance scores at each level in Stage 1 and Stage 2 respectively are shown in bold. As can be seen from the table:
- In Stage 2, our JNT_CW model achieved the highest performance at Rel and State levels.

Particularly at the State level, the F1 score of our JNT_CW is 58.4%, at least 3.5 units higher than those of other systems. This demonstrates the effectiveness of the joint learning method in BEL statement extraction.

- In Stage 1, our JNT_CW model still achieves promising performance, which is close to the rule-based [4,5] and outperforms other systems. This may be due to that the joint model takes into account both relation extraction and function detection, and thus has better noise tolerance to entities.

Table 6. Performance comparison with other models

| Systems | Term | FS | Func | RS | Rel | State |
| --- | --- | --- | --- | --- | --- | --- |
| Rule-based[4,5] | 62.9 | 55.4 | **42.6** | 73.3 | **49.2** | **39.2** |
| Event-based[6] | 34.0 | 10.0 | 8.6 | 25.1 | 41.4 | 20.2 |
| NCU-IISR[8] | 45.0 | 9.5 | 2.7 | 56.7 | 26.4 | 19.7 |
| Att-BiLSTM[11] | 58.6 | 34.3 | 17.7 | 62.3 | 31.6 | 21.3 |
| SBEL-BERT[12] | 59.8 | **59.6** | 28.5 | 72.2 | 40.4 | 30.1 |
| Ours (JNT_CW) | **68.4** | 41.5 | 28.7 | **82.6** | 47.6 | 37.2 |
| Rule-based[4,5] | 82.4 | 56.5 | 30.0 | 82.4 | 65.1 | 25.6 |
| Event-based[6] | 54.3 | 26.1 | 20.8 | 61.5 | 43.7 | 35.2 |
| NCU-IISR[8] | 55.2 | - | - | 63.5 | 44.6 | 33.1 |
| Att-BiLSTM[11] | **97.2** | 34.8 | 26.6 | **96.5** | 65.8 | 46.9 |
| SBEL-BERT[12] | 94.2 | **63.2** | **47.9** | 95.8 | 74.3 | 54.8 |
| Ours (JNT_CW) | 91.8 | 46.5 | 43.1 | 93.5 | **75.0** | **58.4** |

## 5 Discussion and Analysis

This section compares and analyzes Stage 2 performance on the test set from two perspectives of function weighting and joint learning, and further discusses the possible reasons behind the performance improvements.

**5.1 Function weighting**

Table 7 compares the State 2 performance scores for different function types in the separate models when the loss function is weighted or unweighted. Since the sub-tasks of function detection and relation extraction are trained separately, the performance scores are evaluated in terms of function instances. We take the average of five runs as the overall performance, and the highest P/R/F1 scores in each row are shown in bold. As can be seen from the table, when weighted, the precision scores of three and overall types increase at the expense of recall decreases for all types. Although the overall F1 score of function weighting is 4.3 units lower than that of un-weighting, the overall performance of BEL statements is still improved.

In order to analyze the above results, we compared the prediction results of function weighting and un-weighting, and find that false-positive instances of *act* and *complex* function types are significantly reduced when the loss function is weighted, as exemplified as follows:

- In the sentence "ADAM 23 / MDC3, a human disintegrin that promotes cell adhesion via interaction with the alphavbeta3 integrin through an RGD-independent mechanism." (PMID:10749942), the biological process entity "cell adhesion" with None function type is incorrectly predicted as complex function type under the unweighted loss function. This might be because the entity is close to the clue word "interaction" of the complex function type.

Table 7. FD performance broken down to function types

| Function types | Percent (%) | SEP[12] | | | SEP_CW | | |
|---|---|---|---|---|---|---|---|
| | | P | R | F1 | P | R | F1 |
| act | 39.1 | 42.7 | **35.4** | **38.6** | **59.0** | 20.0 | 29.3 |
| complex | 31.9 | 85.8 | **33.6** | **48.2** | **94.2** | 30.7 | 46.0 |
| deg | 8.7 | **66.0** | **46.7** | **54.5** | 60.0 | 30.0 | 39.7 |
| pmod | 7.2 | **6.7** | **4.0** | **5.0** | 0.0 | 0.0 | 0.0 |
| sec | 5.8 | 47.0 | **80.0** | 59.0 | **56.8** | 72.0 | **63.1** |
| tloc | 7.2 | **83.3** | **36.0** | **48.4** | 80.0 | 20.0 | 31.4 |
| FD | 100.0 | 45.5 | **54.5** | **49.5(±5.3)** | **57.4** | 38.0 | 45.2(±2.7) |

- In the sentence "Indeed, SIRT5 null mice fail to upregulate CPS1 activity and show elevated blood ammonia during fasting." (PMID:20157539), the *None* function type of mouse gene "SIRT5" was incorrectly predicted as *act* function type under the unweighted loss function. That might be because according to statistics on the training set, nearly 25% of the entity mentions with the keyword "null" nearby have *act* function type, and the function weighting can avoid such false positive instances.

**5.2 Joint vs. Separate**

It can be seen from Table 4 that the performance discrepancies between the separate models and the joint model is trivial. Therefore, in Table 8, we only compare the Stage 2 performance differences in function detection at the SBEL_FD level. Note that the performance discrepancies of SEP in Table 7 and Table 8 are caused by different evaluation levels. Similarly, the higher P/R/F1 scores in each row among the two models are shown in bold.

Table 8. performance comparison between SEP and JNT at the SBEL_FD level

| Function types | Percent (%) | SEP [12] | | | JNT | | |
|---|---|---|---|---|---|---|---|
| | | P | R | F1 | P | R | F1 |
| act | 36.8 | 50.2 | **36.0** | **41.8** | **52.6** | 33.1 | 40.6 |
| complex | 41.4 | 78.5 | **35.8** | **49.2** | **100.0** | 8.3 | 15.2 |
| deg | 6.9 | **100.0** | **70.0** | **82.2** | **100.0** | 60.0 | 74.7 |
| pmod | 5.7 | 6.7 | 5.0 | 5.7 | **80.3** | **76.0** | **77.1** |
| sec | 4.6 | 35.2 | 75.0 | 47.8 | **66.4** | **85.0** | **74.0** |
| tloc | 4.6 | 43.3 | 40.0 | 40.8 | **60.7** | **48.0** | **52.9** |
| SBEL_FD | 100.0 | 57.7 | 25.8 | 35.6(±5.3) | **66.0** | **29.3** | **40.5(±1.5)** |

It is observed from the table that the reason why the joint model is superior to the separate models in overall P/R/F1 scores is that the former has higher precision in all types than the latter. Although the recall scores of *act*, *complex*, and *deg* function types decrease, those of the other three types are significantly greater than the latter.

By observing and comparing the function instances predicted by the two models, we find that for function types of *pmod*, *sec* and *tloc* with relatively fewer instances, the separate model cannot predict correctly as the joint model can do. The reason may be that the function instance generation on the training set for the separate model ignores possible multiple functions of the same entities, resulting in a large number of entities with function clue words not correctly annotated. This interferes with the training of the function detection model and affects the generalization ability of the model. In the joint model, relation extraction and function detection are jointly trained, and those entity functions derived from the relation label of *None* will not interfere with the training phase, and the generalization ability of the joint model is thus significantly improved.

Taking the *pmod* function type as an example, among the training instances of the separate model for function detection, nearly 50% of the entities with the keyword "phosphorylation" are not labeled as *pmod* function type because they do not participate in any relation. For example, in the sentence "These effects (TLR2,3,5,6 activation of VEGF and IL8) were prevented by treatment with a selective inhibitor of ***EGFR*** phosphorylation (AG-1478), a metalloprotease (MP) inhibitor, a reactive oxygen species (ROS) scavenger, and an NADPH oxidase inhibitor." (PMID:18375743), the protein entity ***EGFR*** that should have *pmod* function does not involve any relation instances with other entities.

## 6 Conclusion and Future work

This paper proposes a joint learning model of relation extraction and function detection based on SBEL statements, which aims to capture the potential relationship between two sub-tasks. Function type weighting is also applied to the joint loss function, so as to improve the precision of function detection and further improve the performance of BEL causal relation extraction. Experimental results show that the joint model mitigates the issue caused by the independence of relation extraction and function detection in the separate models, that is, when the relation between two entities does not exist, the function types of the two entities are forcibly marked as *None*, regardless of the true function types of the entities in the sentence. Because the noise in the training instances of function detection is eliminated, and the interaction between relation and function is fully utilized, joint learning significantly improves the precision and recall of function detection, and thus improves the performance of BEL statements.

Similar to the previous BEL statement extraction work, although the performance at the function level is significantly improved, it is still far from satisfactory. How to adopt more effective methods, including the incorporation of knowledge base and external resources, to improve the performance at the function level is one of our future research directions.

## Funding

This research is supported by the National Natural Science Foundation of China [61976147; 2017YFB1002101] and the research grant of The Hong Kong Polytechnic University Projects [#1-W182].